\pdfoutput=1

\documentclass[11pt]{article}

\usepackage[preprint]{acl}

\usepackage{times}
\usepackage{latexsym}
\usepackage{CJKutf8}
\usepackage[T1]{fontenc}

\usepackage[utf8]{inputenc}

\usepackage{microtype}

\usepackage{inconsolata}

\usepackage{comment}
\usepackage{amsmath}
\usepackage{amssymb}
\usepackage{inconsolata}
\usepackage{graphicx}
\usepackage{subfigure}
\usepackage{caption}
\usepackage{subcaption}
\usepackage{booktabs}
\usepackage{multirow}
\usepackage{arydshln}
\usepackage{amsmath}
\usepackage{cases}
\usepackage{color}
\usepackage{paralist}
\usepackage{bm}
\usepackage{appendix}

\title{Sharing Matters: Analysing Neurons Across Languages and Tasks in LLMs}

\author{%
Weixuan Wang\textsuperscript{1}, Barry Haddow\textsuperscript{1}, Minghao Wu\textsuperscript{2},
Wei Peng\textsuperscript{3}, Alexandra Birch\textsuperscript{1}
 \\[1ex]
\textsuperscript{1}School of Informatics, University of Edinburgh \\
\textsuperscript{2}Monash University \quad \textsuperscript{3}Huawei Technologies Co., Ltd.\\
\texttt{\{weixuan.wang, bhaddow, a.birch\}@ed.ac.uk} \\
\texttt{minghao.wu@monash.edu} \quad \texttt{peng.wei1@huawei.com}
}

\begin{document}

\renewcommand{\tableautorefname}{Table}
\renewcommand{\sectionautorefname}{Section}
\renewcommand{\subsectionautorefname}{Section}
\renewcommand{\subsubsectionautorefname}{Section}
\renewcommand{\figureautorefname}{Figure}
\newcommand{\subfigureautorefname}{Figure}
\newcommand{\linenoautorefname}{Line}
\renewcommand{\appendixautorefname}{Appendix}

\maketitle

\begin{abstract}

Large language models (LLMs) have revolutionized the field of natural language processing (NLP), and recent studies have aimed to understand their underlying mechanisms. However, most of this research is conducted within a monolingual setting, primarily focusing on English. Few studies have attempted to explore the internal workings of LLMs in multilingual settings. In this study, we aim to fill this research gap by examining how neuron activation is shared across tasks and languages. We classify neurons into four distinct categories based on their responses to a specific input across different languages: \textit{all-shared}, \textit{partial-shared}, \textit{specific}, and \textit{non-activated}. Building upon this categorisation, we conduct extensive experiments on three tasks across nine languages using several LLMs and present an in-depth analysis in this work. Our findings reveal that: (i) deactivating the \textit{all-shared neurons} significantly decreases performance; (ii) the shared neurons play a vital role in generating responses, especially for the \textit{all-shared neurons}; (iii) neuron activation patterns are highly sensitive and vary across tasks, LLMs, and languages. These findings shed light on the internal workings of multilingual LLMs and pave the way for future research. We release the code to foster research in this area. \footnote{\url{https://github.com/weixuan-wang123/multilingual-neurons}}

\end{abstract}

\section{Introduction}
\label{sec:introduction}

Large language models (LLMs) have demonstrated remarkable capabilities in recent studies, excelling in both understanding and generating text across various languages \citep{mllm1, mllm2, mllm3}. Despite their proven effectiveness, the intricate mechanisms underlying their processing remain largely opaque. This opacity has given rise to a growing field of research aimed at interpreting the internal workings of the Transformer architecture \citep{circuits,residual}. To enhance interpretability and investigate specific aspects of model behavior, researchers have increasingly focused on the components of these models. Recent studies have explored the role of Feed-Forward Networks (FFNs) within LLMs, proposing that these components function as key-value memories for storing factual and linguistic knowledge \citep{gradient,contribution,key-value}. 
While these studies have analyzed neuron behaviors based on activation states in monolingual settings, there remains a significant gap in our understanding of how neurons behave in multilingual contexts.

\begin{figure}[t]
    \centering
    \includegraphics[scale=0.47]{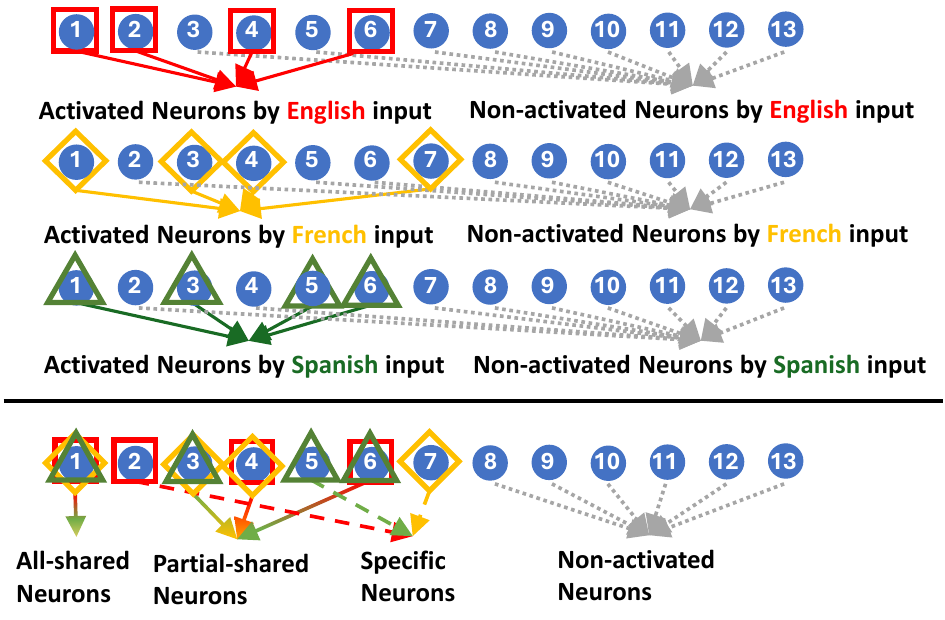}
    \caption{A comparison of neuron analysis with different type designs in multilingual settings with the same semantic input, in which we define four types of neurons in one layer of LLM. }
    \label{fig:4type-neuron}
\end{figure}

To address this research gap, recent research attempts to unveil the mechanistic interpretability of multilingual LLMs. 
\citet{unveil} categorized neurons into two coarse-grained groups: language-agnostic (shared across languages) and language-specific (unique to a language). However this categorization oversimplifies the complexity observed in cross-lingual studies, where neuron overlap varies significantly between languages \citep{same,region,babel}. Additionally, most research has been confined to single-task analyses, overlooking how neuron types might shift across diverse tasks \citep{unveil, specific, mt-neuron}. This underscores the need for a more nuanced, fine-grained classification method to enhance our understanding of the multifaceted roles of neurons in multilingual LLMs.

In this work, our research introduces a fine-grained classification of neurons, enabling a detailed exploration of their functions across languages. For a specific English example and its translations in eight other languages, we categorize neurons into four distinctive types (see \autoref{fig:4type-neuron}): \textit{all-shared neurons}, which remain active for all the inputs regardless of language; \textit{partial-shared neurons}, which are activated only for inputs in certain languages; \textit{specific neurons}, which are activated exclusively for inputs in one language; and \textit{non-activated neurons}, which are not activated for any inputs. We begin by analysing the importance of each neuron type by deactivating them individually. Then we probe their contributions to generating answers using the Generation Impact Score \citep{contribution} and the Correctness Impact Score \citep{dead}. Furthermore, by examining the percentage of neurons in each type, we analyse activation patterns to gain insights into the internal workings of LLMs. We systematically study neuron behaviours across three distinct tasks, including reasoning, fact probing, and question answering, in nine languages. This analysis utilizes diverse model backbones such as \textsc{BLOOMZ-7b}, \textsc{LLaMA2-7b-chat}, \textsc{BLOOM-7b}, and \textsc{XGLM}.

We provide substantial empirical evidence detailing neuron contributions and activation patterns in this study, leading to several significant findings. Here are the main takeaways:

\begin{itemize}

    \item \textbf{\textit{All-shared neurons} have a significant impact on model performance.} We individually deactivate each type of neurons in LLMs and observe substantial performance declines (up to 87.39\%) across tasks (see \autoref{sec:disable}).

    \item \textbf{\textit{All-shared neurons} are crucial in generating responses.} Both the Generation Impact Score and Correctness Impact Score highlight the significance of the shared neurons in the generation process, and the \textit{all-shared neurons} make substantially more contributions compared to other neuron types (see \autoref{sec:contribution}).

    \item \textbf{Neuron activation patterns vary across tasks, LLMs, and languages.} We observe that the patterns of four types of neurons vary across tasks (see \autoref{sec:behaviors_task}) and LLMs (see \autoref{sec:behaviors_llm}). Moreover, our empirical results show that languages from the same language family do not always exhibit a higher degree of neuron sharing compared with languages from distinct language families (see \autoref{sec:behaviors_language}).

\end{itemize}

\section{Related Work}
\label{sec:related}
The black-box nature of LLMs has given rise to an area of research which aims to interpret the internal mechanism of the Transformer architecture \citep{circuits,residual}. More recently, several studies on LLMs have advanced our understanding of how neurons acquire task-specific knowledge. For instance, \citet{key-value, kn,gradient,contribution} investigated how FFN blocks function as key-value memories and proved that factual knowledge is stored in the neurons. Research work on the sparsity of neurons in FFN blocks showed that many neurons are inactive in various tasks~\citep{sparisity,sparser}. \citet{dead} located these ``dead'' neurons in the lower part of the model (close to inputs) in the English scenario. Despite the insights obtained, these studies have focused exclusively on a monolingual setting.

For multilingual neuron analysis, \citet{unveil} explored the neuron sharing between two languages. \citet{specific,mt-neuron,babel,ability} classified neurons in an FFN block to language-specific and language-agnostic based on predefined threshold.
However, the broad classification into two groups is inadequate for detailed multilingual analysis. Additionally, these studies classified neurons based on the single task \citep{mt-neuron,babel}, without considering the potential adaptation of neurons under various languages and semantics brought forth by inputs from various multilingual tasks. We investigate neurons' behaviors across multiple languages and tasks to this end. 

\section{Fine-Grained Neuron Classification}
\label{sec:method}

\begin{table*}[t]
\scriptsize
\centering
\setlength{\tabcolsep}{6pt}
\begin{tabular}{lcccccccccccc}
\toprule
& \multicolumn{3}{c}{\texttt{XNLI}}                  & \multicolumn{3}{c}{\texttt{KE (EN $\rightarrow$ ALL)}} & \multicolumn{3}{c}{\texttt{KE (ALL $\rightarrow$ EN)}} & \multicolumn{3}{c}{\texttt{Fact Probing}}          \\  \cmidrule(rl){2-4}  \cmidrule(rl){5-7}  \cmidrule(rl){8-10} \cmidrule(rl){11-13}
 & pct. & $\mu_{\textrm{acc}}$ & $\Delta_{\textrm{acc}}$    & pct.            & $\mu_{\textrm{acc}}$            & $\Delta_{\textrm{acc}}$            & pct.            & $\mu_{\textrm{acc}}$           & $\Delta_{\textrm{acc}}$              & pct. & $\mu_{\textrm{acc}}$ & $\Delta_{\textrm{acc}}$  \\ \midrule
baseline            & \phantom{0}0.00\%      & 41.99   & \phantom{0}0.00\%            & \phantom{0}0.00\%                 & 38.39              & \phantom{0}0.00\%                      & \phantom{0}0.00\%                 & 41.74             & \phantom{0}0.00\%                      & \phantom{0}0.00\%      & 41.98   & \phantom{0}0.00\%            \\ \hdashline
w/o. all     & \phantom{0}9.92\%      & \phantom{0}9.38    & \textbf{-77.66\%} & \phantom{0}8.71\%                 & \phantom{0}4.84               & \textbf{-87.39\%}           & 10.17\%                & 13.19             & \textbf{-68.40\%}           & \phantom{0}0.28\%      & 21.86   & -50.31\%          \\
w/o. partial & 10.33\%     & 42.65   & \phantom{0}1.57\%            & 13.36\%                & 40.67              & \phantom{0}5.94\%                      & 10.55\%                & 39.59             & -\phantom{0}5.15\%                     & 36.73\%     & 26.86   & -36.02\%          \\
w/o. specific       & \phantom{0}3.14\%      & 42.07   & \phantom{0}0.19\%            & \phantom{0}4.91\%                 & 40.78              & \phantom{0}6.23\%                      & \phantom{0}3.82\%                 & 40.77             & -\phantom{0}2.32\%                     & 16.56\%     & 12.68   & \textbf{-67.41\%} \\
w/o. non-act.  & 76.61\%     & 35.90   & -14.50\%          & 73.22\%                & 21.96              & -42.80\%                    & 75.46\%                & 19.58             & -53.09\%                    & 46.43\%     & 26.68   & -36.45\%          \\ \hdashline
\multirow{4}{*}{w/o. random}      & \phantom{0}5.00\%      & 42.30   & \phantom{0}0.74\%            & \phantom{0}5.00\%                 & 30.98              & -19.30\%                    & \phantom{0}5.00\%                 & 41.29             & -\phantom{0}1.08\%                     & \phantom{0}1.00\%      & 37.86   & -\phantom{0}9.81\%           \\
         & 15.00\%     & 43.13   & \phantom{0}2.71\%            & 15.00\%                & 31.74              & -17.32\%                    & 15.00\%                & 42.14             & \phantom{0}0.96\%                      & 15.00\%     & 35.38   & -15.72\%          \\
         & 25.00\%     & 43.98   & \phantom{0}4.74\%            & 25.00\%                & 32.40              & -15.60\%                    & 25.00\%                & 42.28             & \phantom{0}1.29\%                      & 35.00\%     & 41.78   & -\phantom{0}0.48\%           \\
         & 75.00\%     & 36.58   & -12.88\%          & 75.00\%                & 13.29              & -65.38\%                    & 75.00\%                & 16.50             & -60.47\%                    & 45.00\%     & 17.06   & -59.36\%    \\
\bottomrule
\end{tabular}
\caption{\label{tab:avg-acc-pct-w/o}
The performance on \texttt{XNLI}, \texttt{Cross-lingual KE}, and \texttt{Fact Probing} tasks, using \textsc{BLOOMZ-7b}, when deactivating \textit{all-shared neurons}, \textit{specific neurons}, \textit{partial-shared neurons}, \textit{non-activated neurons}, and random selected neurons, respectively. The largest reductions are highlighted in \textbf{bold}. ``pct.'' indicates the percentage of the deactivated neurons. $\mu_{\textrm{acc}}$ indicates the macro-average accuracy across languages. $\Delta_{\textrm{acc}}$ indicates the macro-average of relative change (\%) in accuracy across languages.}
\end{table*}

In this section, we provide a detailed description of the 4-way neuron classification that we propose. We begin with some background concerning neurons in the FFN block (\autoref{sec:method_background}). Following this, we define the four types of neurons (\autoref{sec:method_definition}).

\subsection{Neurons in FFN Blocks}
\label{sec:method_background}

\label{theory}
A neuron inside the FFNs is defined as a linear transformation of an input representation followed by a non-linear activation \citep{specific}.
Every FFN block at layer $l$ involves two linear transformations separated by a point-wise activation function. Biases are omitted for brevity:
\begin{gather}
    FFN^l(x^l) = Act(W_K^lx^l)W_V^l
    \label{ffn}
\end{gather}
where $W_K^l\in \mathbb{R}^{d \times d_m}$,$ W_V^l \in \mathbb{R}^{d_m \times d}$ are linear parameter matrices, and $Act(\cdot)$ is a non-linear activation function, where rows in $W_K^l$ and columns in $W_V^l$ are viewed as $d$-dimensional keys $k^l$ and values $v^l$, respectively. $d_m$ is the count of neurons. 
And the output of neurons $A^l:= Act(W_K^lx^l)\in \mathbb{R}^{d_m} $ determines the weighting of the corresponding values in $W_V^l$. 

For the $i$-th neuron and corresponding key $k_i^l$, value $v_i^l$ and activation value $A_i^l$, we can express this relationship using the following formulation:
\begin{gather}
    FFN^l(x^l) = \sum_{i=1}^{d_m}Act(x^l \cdot k_i^l)v_i^l = \sum_{i=1}^{d_m}A_i^l v_i^l
    \label{key-value}
\end{gather}
Following \citet{dead,unveil,specific}, we define a neuron as activated when its activation value satisfies $A_i^l > 0$. Conversely, if the activation value is $A_i^l \leq 0$, the neuron is considered deactivated.

\subsection{Definitions of Four Types of Neurons}

\label{sec:method_definition}
In this work, we categorize the neurons into four types based on their activation values and detail the neuron classification in this section.
To ablate the impact of semantic discrepancies across languages, the datasets used in this work are initially in English and then translated into foreign languages (see \autoref{section4-task}), so we can formulate the $s$-th example as $\bm{X}^{s} = \{X^{s}_{p}\}^{P}_{p=1}$, where $p$ indicates the $p$-th language and $P$ is the total number of languages. Given the $s$-th example $\bm{X}^{s}$, the set of \textbf{\textit{all-shared neurons}} at the $l$-th layer can be defined as:
\begin{gather}
    N_{\textrm{all}}^{s,l} := \bigcap_{p}^{P}  \big\{ n^{i}\in N^l :A_{i,p}^{s,l}  > 0 \big\}.
\end{gather}
where $N^{l}$ is the set of all the neurons at the $l$-th layer and $n^{i}$ is the $i$-th neuron in $N^{l}$.
Furthermore, the \textbf{\textit{non-activated neurons}} is the set of neurons whose activation value is less than or equal to zero in all languages, as follows:
\begin{gather}
    N_{\textrm{non}}^{s,l} := \bigcap_{p}^{P}  \big\{ n^{i}\in N^l :A_{i,p}^{s,l}  \leq 0 \big\}.
\end{gather}
Moreover, the \textbf{\textit{specific neurons}} are the neurons only activated in one specific language and not activated in any other languages, defined as follows: 
\begin{equation}
\begin{aligned}
    N_{\textrm{spec}}^{s,l}& :=  
    \bigcup_{p'}^{P} \big\{ \big\{ n^{i}\in N^l :A_{i,p'}^{s,l} > 0 \big\} \\
    & \bigcap_{\substack{p\\p\neq p'}}^{P} \big\{ n^{i}\in N^l :A_{i,p}^{s,l}  \leq 0 \big\}
    \big\}
\end{aligned}
\end{equation}
Lastly, the remaining neurons are \textbf{\textit{partial-shared neurons}} as they are activated by inputs from a subset of languages:
\begin{gather}
    N_{\textrm{part}}^{s,l} := N^{l} \setminus \big\{  N_{\textrm{all}}^{s,l}  \bigcup   N_{\textrm{non}}^{s,l}  \bigcup N_{\textrm{spec}}^{s,l}  \big\}
\end{gather}
Note that, we only examine the activation state of the last token of the input, as that is when the LLM performs the prediction task.

\section{Experimental Setting}

\subsection{Multilingual Tasks}
\label{section4-task}
We perform analysis on neurons in FFN blocks of various LLMs, harnessing their multilingual capabilities in three diverse tasks which consist of multilingual parallel sentences, including \texttt{XNLI} \citep{XNLI}, \texttt{Fact Probing} \citep{mpararel}, and \texttt{Cross-lingual Knowledge Editing (KE)} \citep{remake}. For the \texttt{Cross-lingual KE}, we analyse the LLMs in two setups, including \texttt{EN (Edit) $\rightarrow$ ALL (Test)} and \texttt{ALL (Edit) $\rightarrow$ EN (Test)}. These test sets across languages are translated from the original English test set. More details are described in Appendix C.

These tasks cover nine diverse languages, including English (en), German (de), Spanish (es), French (fr), Russian (ru), Thai (th), Turkish (tr), Vietnamese (vi), and Chinese (zh). Prompts are detailed in Appendix D.

\subsection{Model Backbones}
\label{section4-llms}

We mainly analyse the contributions and activation patterns of neurons in an instruction-finetuned multilingual model \textsc{BLOOMZ-7b}~\citep{bloomz}. We also include the analysis of other multilingual LLMs: \textsc{BLOOM-7b}~\citep{bloom}, \textsc{LLAMA2-7b-chat}~\citep{llama2}, and \textsc{XGLM}~\citep{xglm}. We use one NVIDIA A100 (40G) for all experiments. 
\section{Shared Neurons Are Crucial to Performance}
\label{sec:disable}

\begin{table}[t] \scriptsize
\centering
\setlength{\tabcolsep}{2pt}

\begin{tabular}{lcccccccccc}
\toprule
settings & pct. & en   & de   & es   & fr   & ru   & th   & tr   & vi   & zh   \\ \midrule
\multicolumn{11}{c}{\texttt{XNLI} task} \\  \midrule
baseline     & 0\%& 53.8 & 41.8& 50.3 & 49.0 & 47.6& 40.9& 34.9& 50.5 & 51.1 \\ \hdashline
w/o. all  & \phantom{0}9.9\%  & \textbf{16.7} & \phantom{0}\textbf{3.5} & \textbf{10.1} & \textbf{10.0} & \phantom{0}\textbf{6.6} & \phantom{0}\textbf{9.0} & \phantom{0}\textbf{1.4} & \textbf{12.1} & \textbf{14.5} \\

w/o. partial & 10.3\% & 52.9 & 40.4& 49.7 & 47.6 & 49.2& 40.3& 36.1& 50.0 & 50.0 \\
w/o. specific  &  \phantom{0}3.1\%   & 53.7 & 41.7& 50.3 & 48.9 & 47.4& 40.6& 35.3& 50.4 & 49.3 \\
w/o. non-act.  &  76.7\%   &  36.6&     31.6    &   33.6&  33.4      &   29.5  &    31.3 &  28.3 &  34.5 &  23.5   \\ \hdashline
\multirow{4}{*}{w/o. random}  & 5\%     & 53.2 & 42.2& 50.7 & 48.8 & 47.4& 40.2& 34.5& 50.1 & 50.9 \\
  &  15\% & 53.1 & 41.8& 50.1 & 48.9 & 47.3& 40.8& 33.8& 50.1 & 50.4 \\
  & 25\%  & 52.6 & 41.7& 50.3 & 48.8 & 46.0& 38.8& 36.2& 50.7 & 49.7 \\
 & 75\%  &  36.0&   28.7      &   40.7 &    36.7      &   28.9      &     25.4 &     23.0 &     38.5 &     32.9  \\  \midrule
\multicolumn{11}{c}{\texttt{Cross-lingual KE (EN (Edit) $\rightarrow$ ALL (Test))} task}  \\  \midrule
baseline    & 0\%& 96.2 & 48.8 & 36.9 & 49.5   & 24.6& \phantom{0}6.3 & 38.8 & 49.4 & 33.4\\ \hdashline
w/o. all  & \phantom{0}8.7\%   & \textbf{11.0} & \phantom{0}\textbf{4.9} & \phantom{0}\textbf{6.1} & \phantom{0}\textbf{4.9} &  \phantom{0}1.9 & \phantom{0}\textbf{0.4} & \phantom{0}\textbf{1.5} & \phantom{0}\textbf{6.1} & \phantom{0}\textbf{2.9} \\
w/o. partial & 13.3\% &   90.2& 51.7 & 46.9 & 48.9  &  25.4  &  \phantom{0}5.5& 35.5 & 50.9& 38.4  \\
w/o. specific  &  \phantom{0}4.9\%    & 96.1 & 54.4&   48.7      & 48.9  &   30.4& \phantom{0}6.3 & 37.9& 51.7& 28.5 \\
w/o. non-act. &  73.2\%  &  36.1  &  15.8  &  18.2  &  17.8  &  \phantom{0}1.9  &  \phantom{0}9.8  &  19.9  &  10.6 &  16.3  \\ \hdashline
\multirow{4}{*}{w/o. random}  & 5\%& 96.1 & 46.9 & 36.6 & 40.4 & \phantom{0}0.8 & \phantom{0}4.4 & 28.7 & 40.8 & 10.1 \\
 & 15\% & 94.8 & 47.1 & 36.1 & 39.4 & \phantom{0}0.8  & \phantom{0}4.3 & 28.4 & 40.4 & 11.1 \\
 & 25\% & 91.5 & 46.8 & 36.2 & 38.6 & \phantom{0}1.1  & \phantom{0}4.4 & 27.9 & 40.4 & 12.1 \\
 & 75\% & 11.1 & \phantom{0}5.5 & \phantom{0}9.3 & 11.3 & \phantom{0}\textbf{0.1} & \phantom{0}2.7 & \phantom{0}1.9 & \phantom{0}8.9 & \phantom{0}7.1 \\  \midrule

\multicolumn{11}{c}{\texttt{Cross-lingual KE (ALL (Edit) $\rightarrow$ EN (Test))} task}  \\  \midrule
baseline    & 0\% & 96.2 & 55.1 & 49.2 & 49.5   & 30.6 & \phantom{0}9.2 & 39.3 & 51.7 & 36.6\\ \hdashline
w/o. all     & 10.2\%   & 24.4   & \textbf{19.5}  & \textbf{13.8}  & \textbf{13.1}  & \phantom{0}\textbf{8.0}   & \phantom{0}\textbf{1.4}   & \textbf{14.5}  & 19.9  & \phantom{0}\textbf{7.1}   \\
w/o. partial & 10.5\%  & 85.1   & 51.3  & 47.8  & 48.0  & 25.4  & \phantom{0}5.2   & 35.1  & 51.4  & 36.1 \\
w/o. specific& \phantom{0}3.8\% & 96.1   & 54.4  & 48.7  & 48.9  & 29.7  & \phantom{0}6.3   & 38.0  & 51.8  & 30.0  \\
w/o. non-act.  & 75.5\%  & \textbf{19.2}  & 19.8  & 15.5  & 14.7  & 11.4  & \phantom{0}2.0   & 18.2   & \textbf{12.0 } & \phantom{0}7.5    \\ \hdashline
\multirow{4}{*}{w/o. random}  & \multicolumn{1}{c}{5\%}& 95.6   & 54.2  & 48.7  & 50.2  & 29.9  & \phantom{0}6.5   & 38.5  & 51.3  & 33.0  \\
 & \multicolumn{1}{c}{15\%}   & 93.9   & 56.1  & 49.1  & 49.9  & 29.2  & \phantom{0}6.4   & 38.2  & 51.0  & 32.6  \\
 & \multicolumn{1}{c}{25\%}   & 91.3   & 55.9  & 48.5  & 49.3  & 28.3  & \phantom{0}6.8   & 37.7  & 50.3  & 29.7  \\
 & \multicolumn{1}{c}{75\%}   & 10.2   & 17.0  & 11.7  & 15.7  & \phantom{0}4.7   & \phantom{0}1.1   & \phantom{0}7.8   & 14.8  & \phantom{0}7.0   \\\midrule

\multicolumn{11}{c}{\texttt{Fact Probing} task}  \\  \midrule
baseline   & 0\% & 72.4   & 41.6  & 56.6  & 58.1  & 37.3  & \phantom{0}5.7   & 39.3  & 57.4  & 51.4  \\ \hdashline
w/o. all     & \phantom{0}0.2\% & 43.4   & 12.9  & 34.4  & 22.4  & 11.4  & \phantom{0}\textbf{5.2}   & 15.2  & 34.5  & 29.0  \\
w/o. partial & 36.7\%   & 43.3   & 20.8  & 31.2  & 30.9  & 14.4  & \phantom{0}2.8   & 24.5  & 34.6  & 29.4  \\
w/o. specific& 16.6\%   & \textbf{18.1}   & \phantom{0}\textbf{9.1}  & \textbf{30.1}  & \phantom{0}\textbf{7.7}  & \phantom{0}\textbf{5.7}   & \phantom{0}5.7   & \phantom{0}\textbf{9.4}  & \textbf{12.3} & \textbf{22.1 } \\
w/o. non-act.  & 46.4\%   & 42.5   & 27.6  & 39.9  & 28.1  & \phantom{0}1.7   & \phantom{0}0.0   & 22.9  & 40.2  & 17.5  \\ \hdashline
\multirow{4}{*}{w/o. random}  & \phantom{0}1\%     & 76.4   & 50.6  & 48.6  & 56.0  & \phantom{0}3.2   & \phantom{0}0.0   & 36.2  & 59.5  & 47.1\\
 & 15\% & 71.5   & 48.5  & 45.6  & 63.5  & \phantom{0}4.3   & \phantom{0}0.0   & 22.7  & 44.5  & 38.2  \\
  & 35\%    & 77.3   & 51.4  & 50.3  & 56.3  & \phantom{0}4.6   & \phantom{0}0.0   & 37.1  & 57.5  & 48.3  \\
  & 45\%    & 29.0   & 21.3  & 16.4  & 17.2  & \phantom{0}0.3   & \phantom{0}0.0   & \phantom{0}9.5   & 25.9  & \phantom{0}6.0   \\

\bottomrule
\end{tabular}
\caption{\label{tab:accuracy-w/o}The performance on three tasks, using \textsc{BLOOMZ-7b}, when deactivating \textit{all-shared neurons}, \textit{specific neurons}, \textit{partial-shared neurons}, \textit{non-activated neurons}, and randomly selected neurons, respectively. The largest reductions are highlighted in \textbf{bold}. ``pct.'' indicates the percentage of the deactivated neurons.}
\end{table}

In this section, we explore how different neuron types affect the performance of the \textsc{BLOOMZ-7b} model by selectively deactivating specific groups of neurons. By setting the activation values of these neurons to zero, we assess their impact on the model's output across various tasks. Specifically, we compare the effects of deactivating four distinct types of neurons and include a control group of randomly selected neurons to evaluate their respective contributions to the model performance. Our experiments involve tasks such as \texttt{XNLI}, \texttt{cross-lingual KE}, and \texttt{fact probing}.

\paragraph{\textit{All-shared neurons} play a crucial role in model performance across different tasks.}  
As shown in \autoref{tab:avg-acc-pct-w/o}, we observe that \textit{all-shared neurons} significantly contribute to the model's performance across various tasks. For instance, in the \texttt{Cross-lingual KE (EN (Edit) $\rightarrow$ ALL (Test))} task, deactivating the \textit{all-shared neurons}, which account for only 8.71\% of the total neurons, results in an 87.39\% decrease in accuracy. Moreover, for the \texttt{Fact Probing} task, deactivating the \textit{all-shared neurons}, which constitute only 0.28\% of the total neurons, causes a substantial 50.31\% performance drop. Furthermore, deactivating the \textit{specific neurons}, which account for 16.56\% of the total neurons, leads to the largest performance decline of 67.41\%. In comparison, deactivating a comparable number of random selected neurons typically results in smaller performance drops, suggesting that \textit{all-shared neurons} are crucial to the performance.

\paragraph{Deactivating neurons does not always result in performance declines.}  
Interestingly, we sometimes observe small performance gains when a small number of neurons are deactivated, as shown in \autoref{tab:avg-acc-pct-w/o}, regardless of the neuron type. To explore this phenomenon further, we provide a breakdown of the results by language in \autoref{tab:accuracy-w/o}. Our analysis reveals that only deactivating the \textit{all-shared neurons} consistently leads to a decline in model performance across various tasks and languages. In contrast, deactivating either \textit{partial-shared neurons} or \textit{specific neurons} can occasionally improve performance for certain languages. For example, in the \texttt{Cross-lingual KE (EN (Edit) $\rightarrow$ ALL (Test))} task, we observe substantial performance improvements in German (de), Spanish (es), and Chinese (zh) when the \textit{partial-shared neurons} are deactivated. We hypothesize that this phenomenon stems from knowledge conflicts encoded in the LLM \citep{xu-etal-2024-knowledge-conflicts}. By deactivating certain neurons, these knowledge conflicts may be mitigated, resulting in enhanced performance.

It is important to note that we conduct similar experiments using the \textsc{LLAMA2-7b-chat} and present the results in Appendix E. 
These additional experiments yield observations and conclusions consistent with those using \textsc{BLOOMZ-7b}.

\section{Probing Neuron Contributions}
\label{sec:contribution}

We demonstrate the significant role of neurons shared across languages, particularly the \textit{all-shared neurons}, in generating answers, as discussed in \autoref{sec:disable}. To gain a deeper understanding of the model's behavior, we conduct further analysis using two metrics: \textbf{Generation Impact Score} and \textbf{Correctness Impact Score}. First, we introduce the definitions of these two metrics in \autoref{sec:contribution_theory}. Then, we analyse and quantify the contributions of each type of neuron in \autoref{sec:contribution_generation} and \autoref{sec:contribution_correctness}, respectively.

\subsection{Generation Impact Score and Correctness Impact Score}
\label{sec:contribution_theory}
In this section, we introduce two measures to quantify the contributions of neurons: Generation Impact Score and Correctness Impact Score.

\paragraph{Generation Impact Score}
Inspired by \citet{contribution}, the Generation Impact Score ($GIS$) evaluates the importance of neurons in generating answers. For the $i$-th neuron at $l$-th layer, the $GIS$ is defined as:
\begin{align}
    GIS_i^l :=  \frac{|A_i^l| \; ||v_i^l||}{\sum_{j=1}^{d_m} |A_j^l| \; ||v_j^l||} 
    \label{contribution}
\end{align}
which is the proportion of its weight to the sum of weights of all neurons in the FFN block. $|A_i^l|$ is the absolute value of activation value and $||v_i^l||$ is the L2-norm of value $v_i^l$ (the column in matrix $W_V^l$).

\paragraph{Correctness Impact Score}
Following \citet{contribution} and \citet{dead}, Correctness Impact Score ($CIS$) assesses a neuron's influence on generating the correct answer. 
\begin{align}
    CIS_{i}^{l} = E_r \cdot A_i^l v_i^l
\end{align}
where $E_r$ is the embedding of the correct answer $r$. A larger $CIS_{i}^{l}$ has a higher probability to produce the correct answer $r$, while a negative $CIS_{i}^{l}$ reduces the probability in generating $r$. Detailed descriptions of the neuron projection are provided in Appendix B.

\begin{figure}[t]
\centering          
\subfigbottomskip=-5pt
\subfigure[English test set.]{\label{bloomz-avg-contribution-XNLI}\includegraphics[scale=0.8]{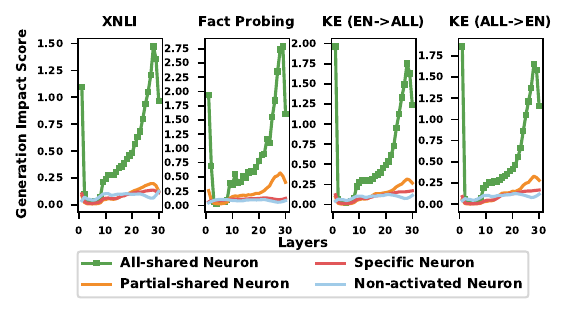}} 
\subfigure[German test set.]{\label{bloomz-avg-contribution-fact}\includegraphics[scale=0.8]{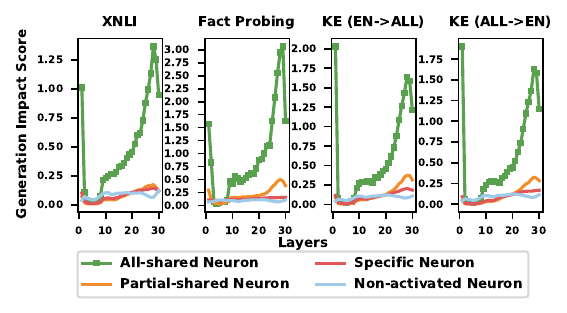}} 
\caption{Average Generation Impact Score of the four types of neurons on the English and German test sets across tasks given by \textsc{BLOOMZ-7b}.}
\label{fig:bloomz-contribution_absolute-allspecific_twolangs} 
\end{figure}

\begin{table*}[t]\scriptsize
\centering
\begin{tabular}{l|cccc|cccc|cccc|cccc}
\toprule
& \multicolumn{4}{c|}{all-shared}  & \multicolumn{4}{c|}{partial-shared}                & \multicolumn{4}{c|}{specific}    & \multicolumn{4}{c}{non-activated}                 \\ \hline
& \multicolumn{1}{c}{max} & \multicolumn{1}{c}{min} & \multicolumn{1}{c}{mean} & var    & \multicolumn{1}{c}{max} & \multicolumn{1}{c}{min} & \multicolumn{1}{c}{mean} & var    & \multicolumn{1}{c}{max} & \multicolumn{1}{c}{min} & \multicolumn{1}{c}{mean} & var    & \multicolumn{1}{c}{max} & \multicolumn{1}{c}{min} & \multicolumn{1}{c}{mean} & var    \\ \hline
en & 1.85  & -0.94 & 0.07   & 0.36   & 0.22  & -0.16 & 1.2e-4 & 1.9e-4 & 0.02  & -0.02 & 2.5e-4 & 3.5e-5 & 0.04  & -0.03 & 2.1e-4 & 5.8e-6 \\
de & 1.03  & -0.60 & 0.02   & 0.07   & 0.13  & -0.13 & 6.7e-5 & 7.1e-5 & 0.07  & -0.03 & 2.3e-5 & 2.3e-5 & 0.02  & -0.01 & 2.9e-6 & 2.9e-6 \\
es & 1.15  & -0.84 & 0.02   & 0.07   & 0.12  & -0.11 & 1.3e-4 & 6.3e-5 & 0.01  & -0.01 & 9.4e-5 & 7.6e-6 & 0.02  & -0.02 & 7.7e-5 & 3.1e-6 \\
fr & 1.06  & -0.78 & 0.01   & 0.05   & 0.15  & -0.11 & 2.1e-4 & 7.8e-5 & 0.03  & -0.04 & 3.6e-5 & 9.9e-6 & 0.02  & -0.02 & 5.6e-5 & 2.8e-6 \\
ru & 0.70  & -0.45 & 3.3e-3 & 5.9e-3 & 0.24  & -0.13 & 2.6e-4 & 7.6e-5 & 0.08  & -0.03 & 1.1e-4 & 1.8e-5 & 0.01  & -0.01 & 3.9e-5 & 1.7e-6 \\
th & 0.50  & -0.90 & 2.6e-3 & 0.02   & 0.17  & -0.10 & 2.2e-4 & 6.8e-5 & 0.03  & -0.05 & 3.4e-5 & 1.8e-5 & 0.01  & -0.01 & 7.1e-5 & 1.9e-6 \\
tr & 0.82  & -0.51 & 0.03   & 0.07   & 0.12  & -0.12 & 1.6e-4 & 7.4e-5 & 0.04  & -0.03 & 6.6e-5 & 1.1e-5 & 0.02  & -0.02 & 9.1e-5 & 3.4e-6 \\
vi & 0.86  & -0.68 & 6.8e-3 & 0.03   & 0.15  & -0.11 & 9.3e-5 & 6.8e-5 & 0.04  & -0.04 & 2.8e-5 & 1.3e-5 & 0.02  & -0.02 & 3.2e-5 & 2.6e-6 \\
zh & 0.52  & -0.42 & 1.9e-3 & 0.02   & 0.17  & -0.20 & 1.5e-4 & 7.7e-5 & 0.08  & -0.07 & 8.5e-5 & 2.6e-5 & 0.02  & -0.01 & 1.7e-5 & 3.1e-6 \\
\bottomrule
\end{tabular}
\caption{\label{tab:max-min-mean-effective-alllangs} Maximum, minimum, average, and variance of Correctness Impact Score of the four types of neurons on the \texttt{Cross-lingual KE (EN (edit) $\rightarrow$ ALL (Test))} task given by \textsc{BLOOMZ-7b}.}
\end{table*}

\paragraph{Comparison}
While both Generation Impact Score ($GIS$) and Correctness Impact Score ($CIS$) measure neuronal influence, they serve different purposes. The $GIS$  quantifies a neuron's overall contribution to the generation process, regardless of output correctness. In contrast $CIS$ specifically measures a neuron's impact on producing accurate responses by incorporating the correct answer's embedding. Thus, the key distinction lies in their consideration of answer correctness: $GIS$ focuses on general generation ability, whereas $CIS$ emphasizes correctness.

\subsection{The Generation Impact of Neuron Types}
\label{sec:contribution_generation}

In this section, we explore the contribution of each neuron type using the Generation Impact Score ($GIS$) described in \autoref{sec:contribution_theory}.

\paragraph{\textit{All-shared neurons} have the greatest impact on generation outputs.}  
As shown in \autoref{fig:bloomz-contribution_absolute-allspecific_twolangs}, we analyse the $GIS$ across layers on the English and German test sets of three tasks (with overall results provided in Appendix F). For both English and German, it can be observed that the \textit{all-shared neurons} almost always achieve the highest $GIS$ across all layers, indicating their significant influence on the model's output generation. The \textit{partial-shared neurons} are the second most influential, particularly in the upper layers. Notably, there is a decrease in the influence of \textit{all-shared neurons} between layers 5 and 10. This can be attributed to the fact that $GIS$ assesses the impact on generating answers, while the lower layers are primarily responsible for input understanding \citep{handle}. Consequently, all types of neurons exhibit lower $GIS$ in these layers. Moreover, previous studies have demonstrated that higher layers capture more abstract, high-level information essential for generation \citep{higher}. These findings suggest that shared neurons play a more significant role in the model's generation capabilities.

\subsection{The Correctness Impact of Neuron Types}
\label{sec:contribution_correctness}

In this section, we assess the effectiveness of each neuron type using the Correctness Impact Score ($CIS$) described in \autoref{sec:contribution_theory}.

\paragraph{\textit{All-shared neurons} have the greatest impact on generating \textit{correct} answers.} In the \texttt{Cross-lingual KE (EN (Edit) $\rightarrow$ ALL (Test))} task, we present the maximum, minimum, average, and variance of $CIS$ for each neuron type across all layers of the \textsc{BLOOMZ-7b} model, as shown in \autoref{tab:max-min-mean-effective-alllangs}. The results reveal that \textit{all-shared neurons} have both the highest maximum and the lowest minimum $CIS$ values, indicating that they have strong impact on generating correct outputs. While \textit{all-shared} and \textit{partial-shared neurons} display a wide variance in $CIS$ (e.g., 1.85 vs. -0.94 and 0.22 vs. -0.16 in English, respectively), \textit{specific neurons} and \textit{non-activated neurons} exhibit much narrower score ranges (approximately $\pm$ 0.07). Furthermore, the \textit{all-shared neurons} also exhibit the largest mean and variance of $CIS$ among all kinds of neurons. 

In conclusion, these findings presented in \autoref{sec:contribution_generation} and \autoref{sec:contribution_correctness} demonstrate that \textit{all-shared neurons} also have the greatest impact on generating both answers and correct answers, highlighting their importance in the model's performance across different languages and tasks.

\begin{figure*}[t]
    \centering
    \includegraphics[scale=0.7]{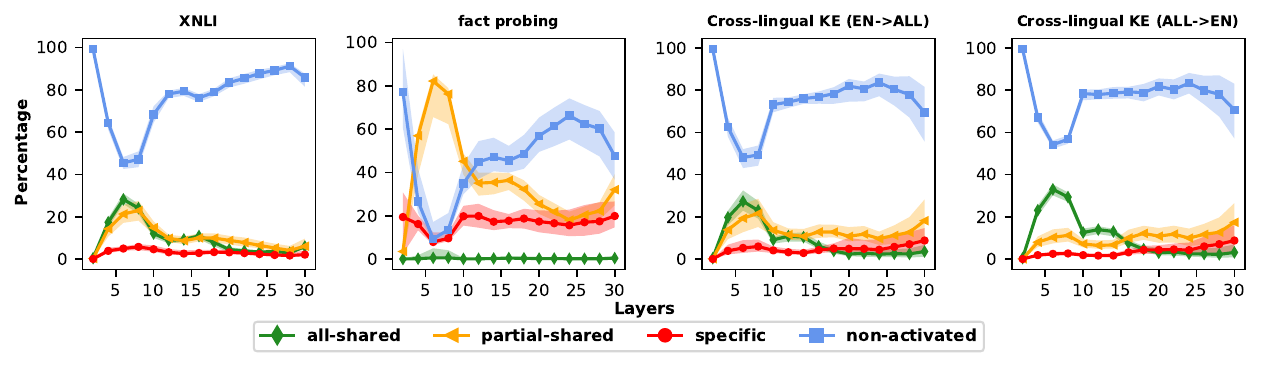}
    \vspace{-0.6em}
    \caption{Neuron activation pattern ($R^{l}_{\{\cdot\}}$) in the \texttt{XNLI}, \texttt{Fact Probing}, \texttt{Cross-lingual KE (EN (Edit) $\rightarrow$ ALL (Test)}, and \texttt{Cross-lingual KE (ALL (Edit) $\rightarrow$ EN (Test)} tasks with \textsc{BLOOMZ-7b} backbone. It shows the percentage of each type of neuron relative to the total number of neurons across layers.}
    \label{fig:bloomz-count-4tasks}
\end{figure*}
\section{Understanding Neuron Activations}
\label{sec:behaviors}

We demonstrate in \autoref{sec:disable} that shared neurons have a significant impact on model performance and investigate their influence on the generation process in \autoref{sec:contribution}. However, the inner patterns of neurons across layers remain unexplored. In this section, we firstly introduce the measure of quantifying neuron activation in \autoref{sec:method_testset}, and then we further illustrate how neuron activation patterns vary across tasks (\autoref{sec:behaviors_task}), LLMs (\autoref{sec:behaviors_llm}) and languages (\autoref{sec:behaviors_language}).

\subsection{Measuring Neuron Activation}
\label{sec:method_testset}
In this section, we explain how to quantify neuron activation patterns based on the definitions in \autoref{sec:method_definition}. Specifically, we measure the percentage of each type of neuron relative to the total number of neurons. Given the $s$-th test instance, the percentage of each neuron type $R^{s,l}_{\{\cdot\}}$ at the $l$-th layer can be defined as follows:
\begin{align}
    R^{s,l}_{\{\cdot\}} = 100 \times \frac{|N^{s,l}_{\{\cdot\}}|}{|N^l|},
\end{align}
where $|\cdot|$ denotes the number of elements in the set. Consequently, the aggregated neuron activation pattern at the $l$-th layer for one dataset containing $S$ instances can be defined as:
\begin{align}
    R^{l}_{\{\cdot\}} = \frac{1}{S} \sum_{s=1}^{S} R^{s,l}_{\{\cdot\}}.
\end{align}

\subsection{Neuron Activations Across Tasks}
\label{sec:behaviors_task}

\paragraph{Neuron activations are task-related.} 
As shown in \autoref{fig:bloomz-count-4tasks}, \textit{non-activated neurons} are typically more prevalent than other types of neurons, except in the \texttt{Fact Probing} task. In this task, there are more \textit{partial-shared neurons} and \textit{specific neurons}, with a negligible amount of \textit{all-shared neurons}, whereas other tasks involve far more \textit{all-shared neurons}. Referring to \autoref{tab:avg-acc-pct-w/o}, deactivating the \textit{specific neurons} and \textit{all-shared neurons} results in the largest and second largest performance declines. These findings demonstrate that the some factual knowledges in LLMs are language-specific and minimally shared across languages, while others are universally shared. We leave more in-depth investigation to the future work.

\paragraph{Neuron sharing peaks at early layers for universal features, declining later for specific ones.}
We present the percentage of each neuron type at each layer in \autoref{fig:bloomz-count-4tasks}. The number of \textit{all-shared neurons} and \textit{partial-shared neurons} typically peaks between the 5th and 10th layers and then gradually decreases in subsequent layers. This trend can be explained by the functional roles of different layers in the model. The initial layers, which are closer to the input data, primarily focus on capturing low-level features such as basic lexical and syntactic patterns. As the network progresses to the later layers (between the 5th and 10th layers), it begins to learn abstract concepts that are relatively universal across different tasks and languages. This universality leads to a higher number of shared neurons in these layers. In contrast, the higher layers specialize in task-specific features and nuances unique to each task, resulting in a decline in neuron sharing. These findings highlight the importance of neuron sharing in LLMs, as shared neurons in the early layers facilitate the transfer of universal knowledge across tasks and languages. They also align with previous research \citep{DBLP:conf/nips/YosinskiCBL14,de-vries-etal-2020-whats,handle,higher}.

\begin{figure}[t]
    \centering
    \includegraphics[scale=0.8]{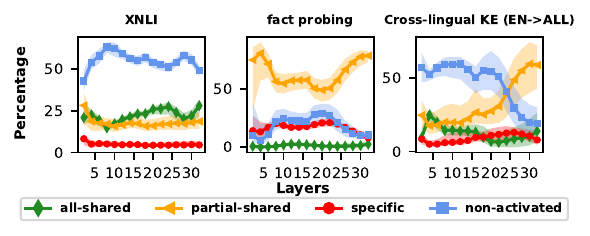}
    \vspace{-0.6em}
    \caption{Neuron activation patterns in the \texttt{XNLI}, \texttt{Fact Probing}, \texttt{Cross-lingual KE (EN (Edit) $\rightarrow$ ALL (Test)} tasks with \textsc{LLaMA2-7b-chat} backbone. }
    \label{fig:llama-3task}
\end{figure}

\begin{figure}[t]
    \centering
    \includegraphics[scale=0.63]{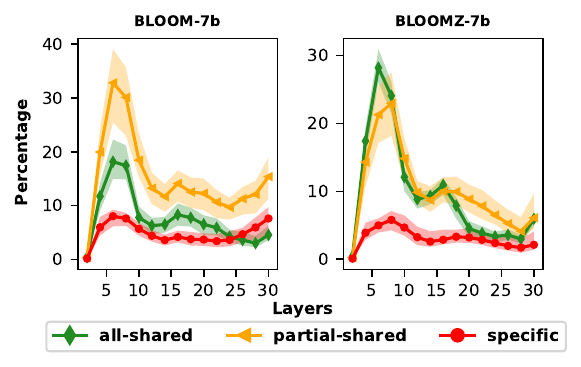}
    \vspace{-0.6em}
    \caption{Comparison of neuron activations with foundation LLM \textsc{BLOOM-7b} (left) and instruction finetuned LLM \textsc{BLOOMZ-7b} (right).}
    \label{fig:bloom-3task}
\end{figure}

\subsection{Neuron Activations Across LLMs}
\label{sec:behaviors_llm}

\paragraph{Different LLMs exhibit different neuron activation patterns.}
To investigate whether neuron activation patterns vary across different multilingual LLMs, we present additional results from \textsc{LLaMA2-7b-chat} in \autoref{fig:llama-3task}. Our analysis reveals that the activation patterns in \textsc{LLaMA2-7b-chat} differ significantly from those observed in \textsc{BLOOMZ-7b}, highlighting the variability across models. Notably, \textsc{LLaMA2-7b-chat} demonstrates a higher degree of neuron sharing, particularly for \textit{partial-shared neurons}. This phenomenon can be attributed to the English-centric nature of \textsc{LLaMA2-7b-chat}. When processing multilingual inputs, the model heavily relies on knowledge transfer from English to other languages, resulting in a substantial number of \textit{partial-shared neurons}. We also present additional results using \textsc{XGLM} \citep{xglm} in Appendix G, aligning with our observations.

\paragraph{Instruction finetuned LLMs exhibit larger proportion of the \textit{all-shared neurons}.}
We conduct additional experiments using the foundation model \textsc{BLOOM-7b} to explore the impact of instruction finetuning on neuron activation patterns. As shown in \autoref{fig:bloom-3task}, the instruction-finetuned \textsc{BLOOMZ-7b} demonstrates a higher percentage of \textit{all-shared neurons} compared to \textsc{BLOOM-7b}. This observation suggests that instruction finetuning may encourage neuron sharing within LLMs, potentially aligning their internal representations across languages. Therefore, instruction-finetuned LLMs, such as \textsc{BLOOMZ-7b}, generally outperform their foundational counterparts.

\subsection{Neuron Activations Across Languages}
\label{sec:behaviors_language}
\begin{figure}[t]
    \centering
    \includegraphics[scale=0.63]{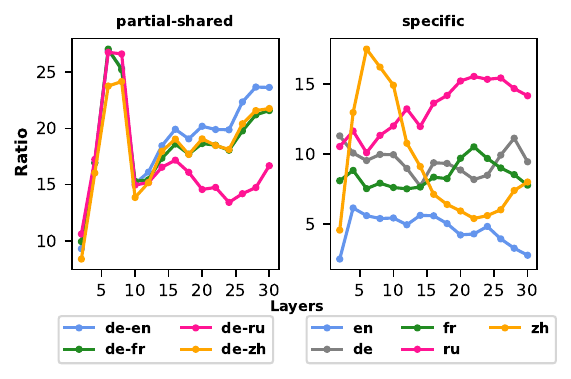}
    \vspace{-0.6em}
    \caption{Neuron activation pattern across languages in the \texttt{Fact Probing} task with \textsc{BLOOMZ-7b} backbone. Left: The ratio of \textit{partial-shared neurons} representing \{en, fr, ru, zh\} shared with German (de). Right: The percentage of \{en, de, fr, ru, zh\} in \textit{specific neurons}.}
    \label{fig:language-partial-specific-factprob}
\end{figure}

\paragraph{Neuron sharing does not completely align with language similarity.} 
We investigate the relationship between language similarity and neuron sharing by analysing the proportion of \textit{partial-shared neurons} for language pairs involving German and several other languages on the \texttt{Fact Probing} task. As shown in \autoref{fig:language-partial-specific-factprob}, our findings reveal that similar languages (e.g., German and French) do not always exhibit higher levels of neuron sharing. For instance, the proportion of \textit{partial-shared neurons} between German and Chinese is nearly identical to that between German and French, despite German and French both belonging to the Indo-European language family, while Chinese belongs to the Sino-Tibetan language family. Furthermore, we observe no consistent pattern in the percentage of \textit{specific neurons} across the languages studied, suggesting that neuron specialization may not directly correlate with language similarity. We leave further exploration of this phenomenon to future work. Additional results on \texttt{XNLI} task are in Appendix I.

Furthermore, we conduct ablation studies to investigate the impact of two key factors on the neuron activation patterns: the size of the backbone model with 0.56b, 1b, 3b, 7b parameters (Appendix J.).

\section{Conclusion}

In this study, we explored the complex mechanisms of neuron activation within multilingual LLMs, addressing the significant research gap in understanding these models beyond a monolingual context. We developed a fine-grained classification for analysing how neurons respond to different tasks and languages. We categorized neurons into four distinct groups: \textit{all-shared}, \textit{partial-shared}, \textit{specific}, and \textit{non-activated}. Our research revealed that neurons shared across all languages proved essential for generating accurate responses, highlighting their pivotal role in multilingual processing. Furthermore, we demonstrate that neuron sharing is task-related, and, it does not always align with language similarity. Our study improves the understanding of the internal workings of multilingual LLMs and fosters future research in this direction.
\section*{Acknowledgement}

This work is funded by EU Horizon Europe (HE) Research and Innovation programme grant No 101070631, and UK Research and Innovation under the UK HE funding grant No 10039436.

The computations described in this research were performed using the Baskerville Tier 2 HPC service (https://www.baskerville.ac.uk/). Baskerville was funded by the EPSRC and UKRI through the World Class Labs scheme (EP/T022221/1) and the Digital Research Infrastructure programme (EP/W032244/1) and is operated by Advanced Research Computing at the University of Birmingham.

\bibliography{custom}

\clearpage
\appendix

\section{Detailed Interpretation of Projection in Vocabulary Space}
\label{appendix:theory}
There is a residual connection in the each layer of transformer, where the hidden state is:

\begin{gather}
    h^l = x^l + FFN^l(x^l)
\end{gather}

In order to analyze the attribution of neurons, we explore how the output distribution in the vocabulary space changes when the representation $x^l$ (before the FFN update) is added with the output of neurons $A_i^lv_i^l$. With the embedding matrix $E$, we map each vector into the vocabulary space $\nu$. For each token $w$, the probability is calculate with the softmax function:

\begin{equation}
\begin{aligned}
&p(w|x^l+A_i^lv_i^l,E) \\  &= \frac{exp(E_w \cdot x^l + E_w \cdot A_i^lv_i^l)}{Z(E(x^l+A_i^lv_i^l))} \\ & \propto  exp(E_w \cdot x^l) \cdot exp(E_w \cdot A_i^lv_i^l)
\end{aligned}
\end{equation}
where $E_w$ is the embedding of $w$, and $Z(\cdot)$ is the constant softmax normalization factor. The $E_w \cdot x^l$ can be viewed as a static score of $w$ that is independent of the input to the model. Thus, the projection $E_w \cdot A_i^lv_i^l$ induces a ranking over the vocabulary. So we use the projection as effective score to detect the responsibility of neurons.

\section{Tasks}
\label{appendix:task}
\begin{itemize}
    \item \texttt{XNLI.} Natural Language Inference 
    is a multilingual natural languages inference dataset, containing 5000 items. Each test sample consists of a premise and a hypothesis, requiring an LLM to determine whether a hypothesis is entailed, contradicted, or neutral conditioned on the premise. 
    \item \texttt{\texttt{Fact Probing}}. LLMs are used to predict factual answers in response to corresponding probing prompts. A multilingual factual knowledge dataset (mParaRel) capturing 38 binary relations (e.g., \textit{X born-in Y}) is used in the analysis. We seletc the relation of ``capital'' subset (\textit{X capital Y}) as testset, including 348 items.
    \item \texttt{Cross-lingual Knowledge Editing (KE)}. MzsREis a multilingual question-answering dataset, containing 743 items for each language.
    It provides co unterfactual edited knowledge in the context and requires an LLM to produce the corresponding answer according to the context. We evaluate LLMs in two Cross-lingual KE scenarios: 1) EN (Edit) $\rightarrow$ ALL (Test): edit in English and test in other languages and 2) ALL (Edit) $\rightarrow$ EN (Test): edit in other languages and test in English.
\end{itemize}

\section{Prompts}
\label{appendix:prompt}
For the \texttt{Fact Probing} task, we use the P36 sub-testset, which describe facts of entities in a relation of ``capital''. The prompt is framed as `` The capital of \{X\} is '' where ``\{X\}'' is the subject (sovereign state) and LLMs are required to predict the object (capital city). We keep at least three paraphrase prompts from mParaRel for each language to ensure a level of diversity. 

For the Natural Language Inference (\texttt{XNLI}) task, we frame the prompt as `` Take the following as truth: \{premise\} Then the following statement: `\{hypothesis\}' is `true', `false', or `inconclusive'? ''

For the \texttt{Cross-lingual KE} task, we format the prompt as `` \{context\} Question: \{question\} Answer: ''. The same language is used for the questions and the answers, but the context is in a different language.

\section{Supplemental Results on Deactivating Neurons}
\label{appendix:llama-accuracy}

\begin{table}[t] \scriptsize
\centering
\setlength{\tabcolsep}{1.5pt}
\begin{tabular}{lcccccccccc}
\toprule
settings & pct. & en   & de   & es   & fr   & ru   & th   & tr   & vi   & zh   \\ \midrule
baseline   & 0\%         & 59.1          & 47.6         & 50.1          & 47.0          & 49.1         & 41.4         & 40.2         & 51.6          & 46.1          \\  \hdashline
w/o. all-shared  &  22.42\%  & \phantom{0}\textbf{3.0}  & \phantom{0}\textbf{3.6} & \phantom{0}\textbf{4.4}  & \phantom{0}\textbf{1.9}  & \phantom{0}\textbf{4.7} & \phantom{0}\textbf{6.9} & \phantom{0}\textbf{3.6} & \textbf{13.5} & \phantom{0}\textbf{4.8}  \\
w/o. partial-shared & 17.48\% & 59.1          & 48.4         & 51.5          & 47.9          & 49.7         & 42.9         & 41.5         & 50.8          & 48.0          \\
w/o. specific  &   4.75\%   & 59.2          & 47.3         & 49.9          & 47.0          & 49.1         & 41.9         & 40.1         & 51.4          & 46.2          \\
w/o. non-activated  &  55.35\%   &   30.5 & 13.8 & 12.0 & 11.9 & 12.4  &  5.0      &   14.2     &   13.4      &    5.2       \\ \hdashline
\multirow{4}{*}{w/o. random} & 5\%     & 58.7          & 47.7         & 50.2          & 48.2          & 49.0         & 41.7         & 40.0         & 49.9          & 45.7          \\
 &  15\%  & 52.7          & 44.6         & 47.2          & 46.4          & 44.5         & 38.4         & 40.1         & 48.6          & 45.2          \\
&  25\%  & 46.1          & 42.4         & 41.3          & 43.3          & 40.1         & 34.5         & 39.7         & 38.7          & 40.7          \\
&  55\% & 28.7& 30.2& 28.6& 30.3& 25.8& 19.0& 27.1& 28.2 & 25.0 \\
\bottomrule
\end{tabular}
\caption{\label{tab:accuracy-w/o-llama}The accuracy in \texttt{XNLI} task with \textsc{LLAMA2-7b-chat} backbone when deactivating four types of neurons.} 
\end{table}

In order to further prove the importance of \textit{all-shared neurons} across LLMs, we conduct the experiments with deactivating neurons on the \texttt{XNLI} task with \textsc{LLAMA2-7b-chat} backbone. The results in \autoref{tab:accuracy-w/o-llama} show that there is more significant decline when \textit{all-shared neurons} are deactivated. It demonstrates that \textit{all-shared neurons} play a key role in predicting correct answers across LLMs.

\section{Generation Impact Score of Different Tasks}
\label{appendix:contribution}

\begin{figure*}[ht]
    \centering
    \includegraphics[scale=0.2]{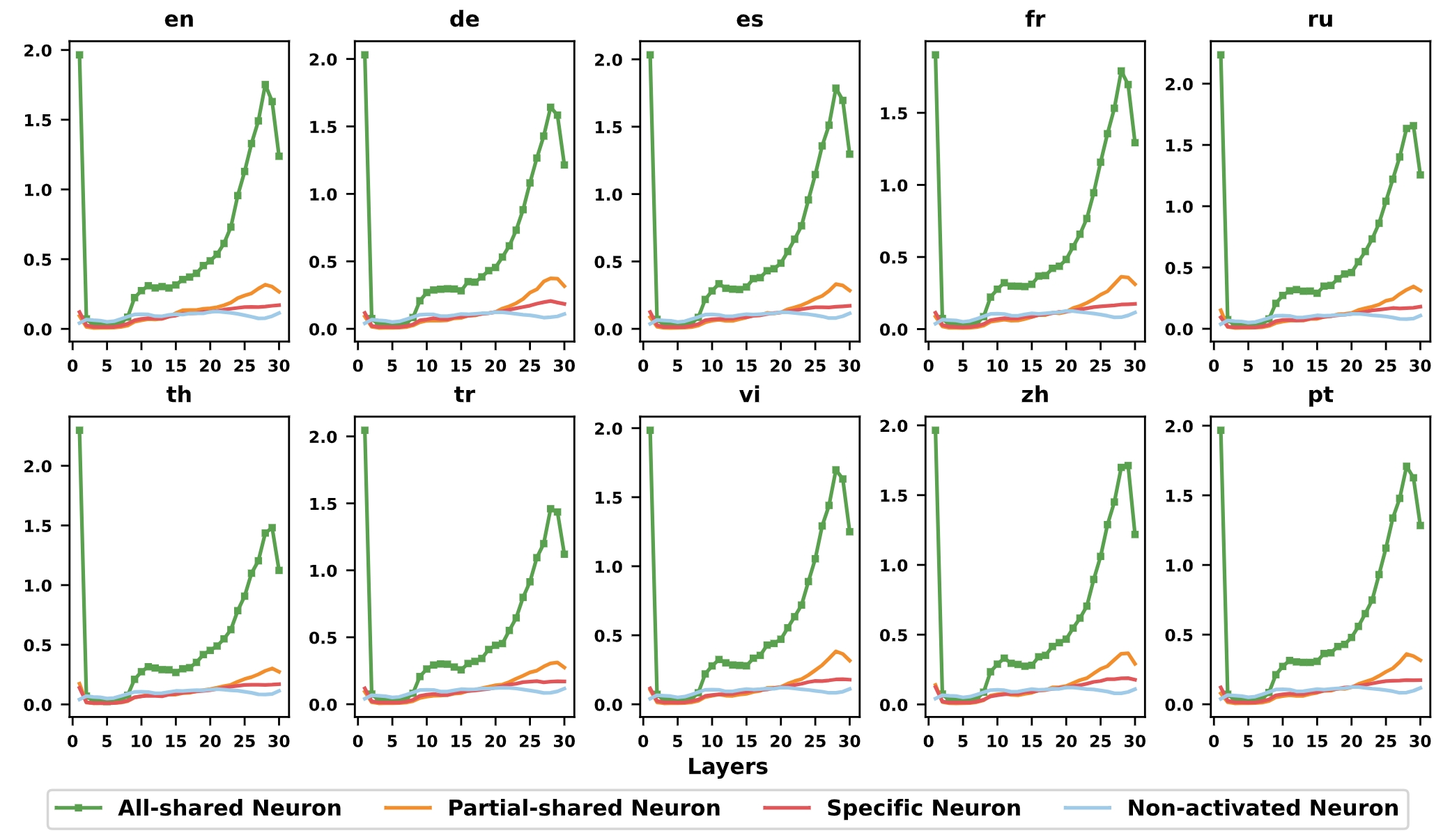}
    \caption{Generation Impact Score on the \texttt{Cross-lingual KE (EN (edit) $\rightarrow$ ALL (Test))} task with \textsc{BLOOMZ-7b} backbone.}
    \label{fig:bloomz-contribution-en2xx}
\end{figure*}

\begin{figure*}[ht]
    \centering
    \includegraphics[scale=0.2]{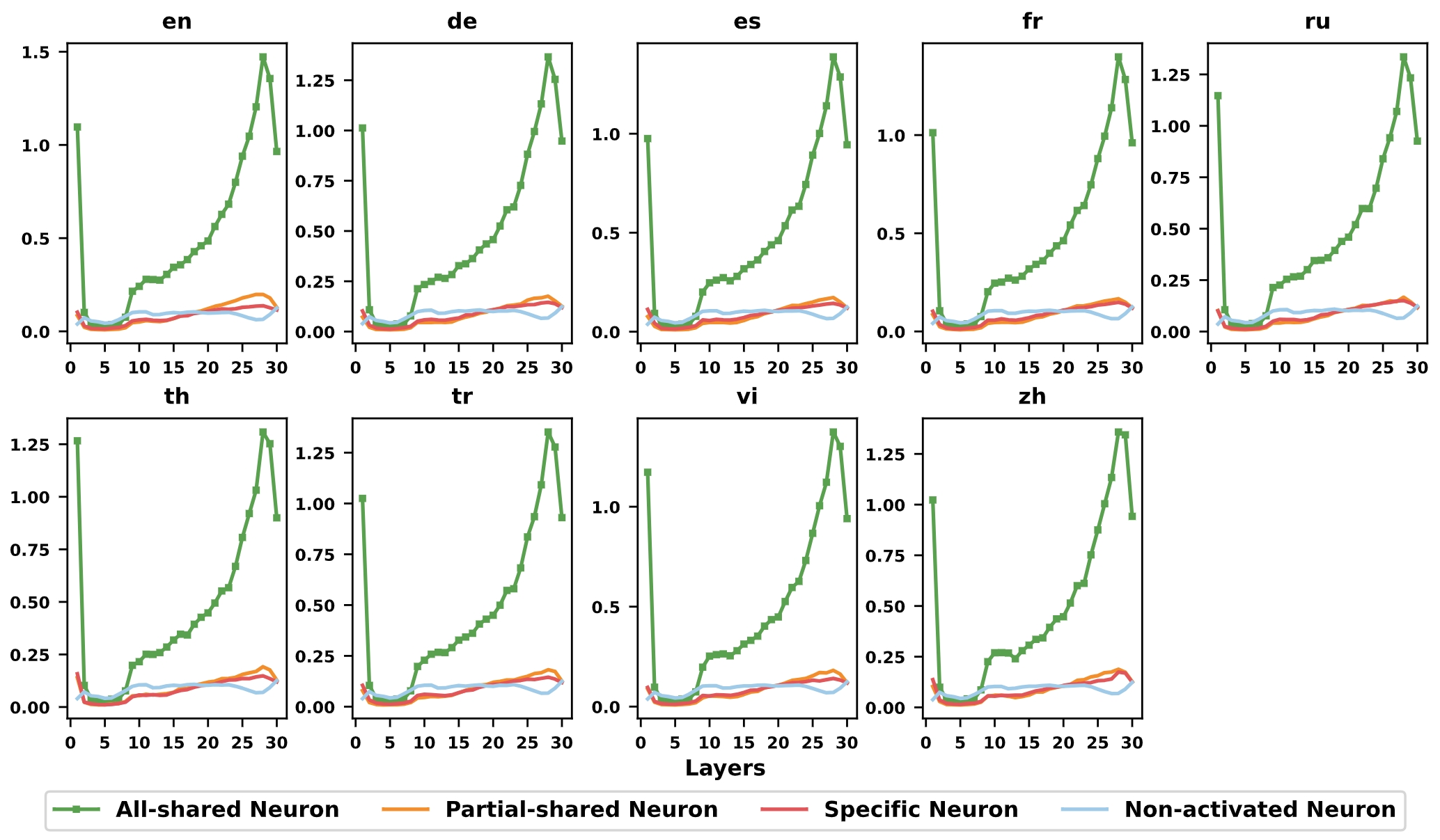}
    \caption{Generation Impact Score on the \texttt{XNLI} task with \textsc{BLOOMZ-7b} backbone.}
    \label{fig:bloomz-contribution-xnli}
\end{figure*}

The Generation Impact Score of the four types of neurons evaluated on the \texttt{Cross-lingual KE (EN (edit) $\rightarrow$ ALL (Test))} and \texttt{XNLI} tasks across languages are shown in \autoref{fig:bloomz-contribution-en2xx} and \autoref{fig:bloomz-contribution-xnli}.

\section{Supplemental Results on Neurons Activation Patterns across LLMs}
\label{appendix:LLMs}

\begin{figure*}[ht]
    \centering
    \includegraphics[scale=0.7]{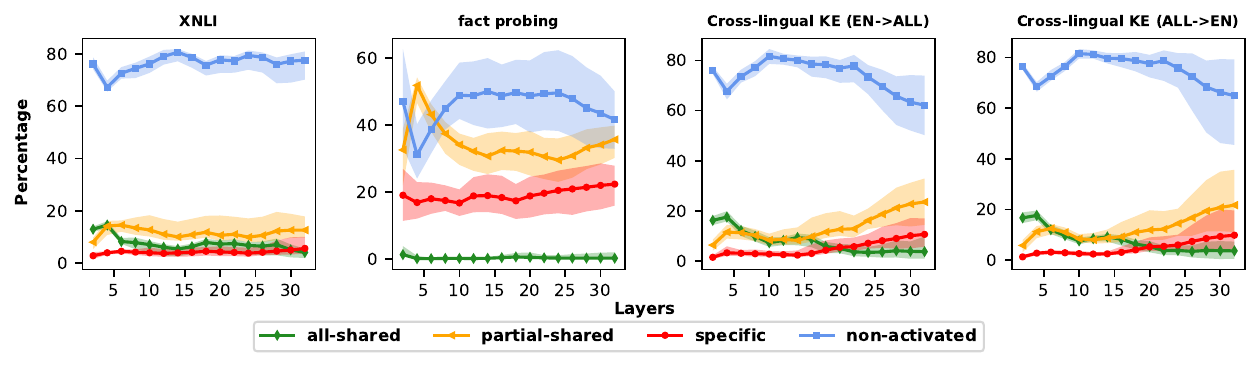}
    \caption{Neuron activation pattern in \texttt{XNLI}, \texttt{Fact Probing}, and \texttt{Cross-lingual KE} tasks with \textsc{XGLM} backbone.}
    \label{fig:xglm-count-4tasks}
\end{figure*}

We further study the neuron activation patterns in another multilingual LLM (\textsc{XGLM}). The results of \textsc{XGLM} backbone are captured in \autoref{fig:xglm-count-4tasks}.

\section{Supplemental Results on Neurons Activation Patterns of Foundation LLM \textsc{BLOOM-7b}}
\label{appendix:bloom}

\begin{figure}[ht]
    \centering
    \includegraphics[scale=0.7]{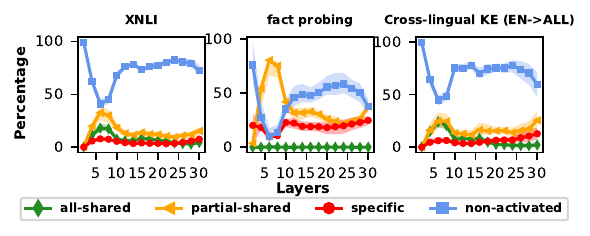}
    \caption{Neuron activation pattern in \texttt{XNLI}, \texttt{Fact Probing}, and \texttt{Cross-lingual KE} tasks with \textsc{BLOOM-7b} backbone.}
    \label{fig:bloom-appendix}
\end{figure}

We further explore the neuron activation patterns across various tasks in the foundation LLM (\textsc{BLOOM-7b}). The results of \textsc{BLOOM-7b} backbone are captured in \autoref{fig:bloom-appendix}.

\section{Neuron Activation Across Languages on \texttt{XNLI} Task}
\label{appendix:language-characteristics}

\begin{figure}[ht]
    \centering
    \includegraphics[scale=0.7]{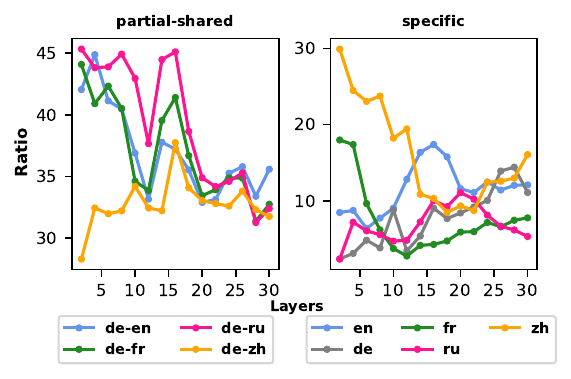}
    \vspace{-0.6em}
    \caption{Aggregated neuron activation pattern across
languages in the \texttt{XNLI} task. Left: The ratio of partially-shared neurons representing \{en, fr, ru, vi\} shared with German (de). Right: The percentage of \{en, de, fr, ru, vi\} in specific neurons.}
    \label{language-partial-specific-XNLI}
\end{figure}

We analyze the shared proportion of German with other languages in \textit{partial-shared neurons} and the \textit{specific neuron} ratios for each language derived from the \texttt{XNLI} task in Figure~\ref{language-partial-specific-XNLI}. The shared ratio of German with Russian (in different language family) is higher than the ratio of German with French (in the same language family), confirming the conclusion in Section 7.4.

\section{Influence of Model Scale}
\label{appendix:scale}
\begin{figure}[ht]
    \centering
    \includegraphics[scale=0.78]{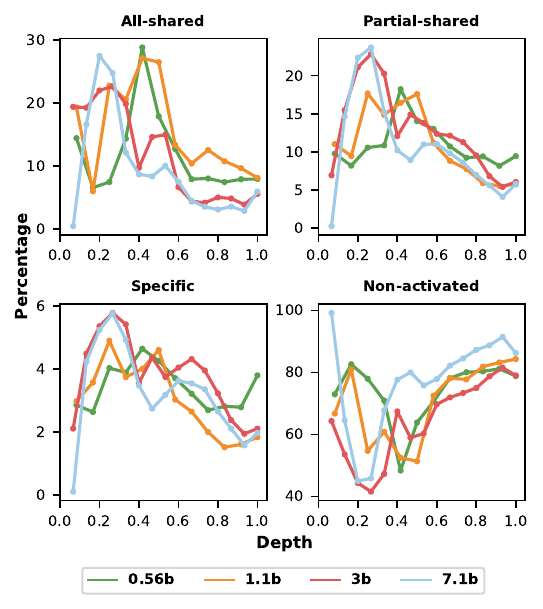}
    \caption{Neuron activation patterns in a \texttt{XNLI} task with the \textsc{BLOOMZ} size as 0.56b, 1b, 3b, 7b.}
    \label{bloom-XNLI-scale}
\end{figure}

We investigate neuron activation patterns across the \textsc{BLOOMZ} series with 0.56b, 1b, 3b, 7b parameters in a \texttt{XNLI} task. As shown in the results captured in Figure~\ref{bloom-XNLI-scale}, no identifiable pattern difference can be observed to indicate a scale law effect. However, the scale of the model is limited, potentially leading to unreliable results in this experiment. More \textit{non-activated neurons} in the upper layers of \textsc{BLOOMZ-7b} may reflect on a higher level of sparsity for a larger LLM. 
\section{Neuron Activation Patterns in Few-shot In-context Learning}
\label{appendix：context}
\begin{figure}[ht]
    \centering
    \includegraphics[scale=0.78]{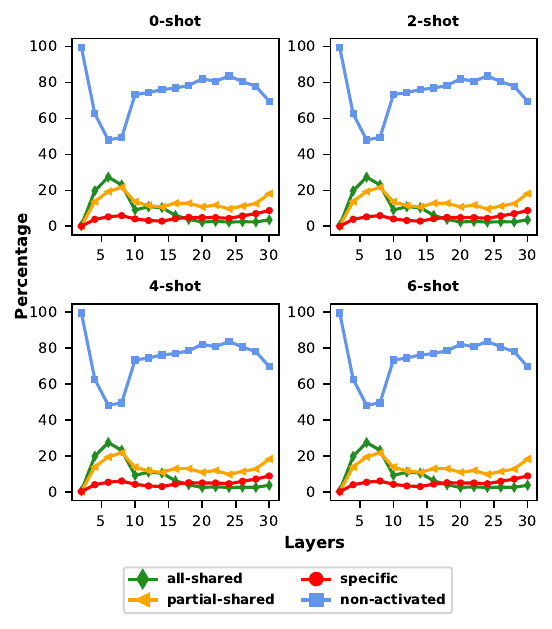}
    \caption{Neuron activation patterns in \texttt{Cross-lingual KE (EN (edit) $\rightarrow$ ALL (Test))} task with \textsc{BLOOMZ-7b} backbone under the in-context learning.}
    \label{bloomz-en2xx-fewshot}
\end{figure}

In-context learning (ICL) can improve the performance of an LLM under the guidance of few-shot examples in a \texttt{Cross-lingual KE} task. We further explore the impact of few-shot examples on neuron activation patterns. We compare the results of an LLM with 0-shot, 2-shot, 4-shot, 6-shot examples in a \texttt{Cross-lingual KE (EN (edit) $\rightarrow$ ALL (Test))} task. Four types of neurons in scope have almost identical activation patterns across various few-shot examples (Figure~\ref{bloomz-en2xx-fewshot}). Although in-context examples lead to no observable neuron activation pattern changes, more examples lead to better performances. Could ICL lead to a better neuron activation composition instead of invoking more neurons? We leave this to a future study.

\end{document}